# Initial Investigation of LLM-Assisted Development of Rule-Based Clinical NLP System

Jianlin Shi, MD, PhD[1,2], Brian T. Bucher, MD, MS[1]
[1]University of Utah, Salt Lake City, UT, USA; [2]VA Salt Lake City Health Care System, UT

**Abstract**
*Despite advances in machine learning (ML) and large language models (LLMs), rule-based natural language processing (NLP) systems remain active in clinical settings due to their interpretability and operational efficiency. However, their manual development and maintenance are labor-intensive, particularly in tasks with large linguistic variability. To overcome these limitations, we proposed a novel approach employing LLMs solely during the rule-based systems development phase. We conducted the initial experiments focusing on the first two steps of developing a rule-based NLP pipeline: find relevant snippets from the clinical note; extract informative keywords from the snippets for the rule-based named entity recognition (NER) component. Our experiments demonstrated exceptional recall in identifying clinically relevant text snippets (Deepseek: 0.98, Qwen: 0.99) and 1.0 in extracting key terms for NER. This study sheds light on a promising new direction for NLP development, enabling semi-automated or automated development of rule-based systems with significantly faster, more cost-effective, and transparent execution compared with deep learning model-based solutions.*

**Introduction**
Rule-based natural language processing (NLP) systems have been widely used in clinical applications, especially in the industry field[1], despite significant advances in deep learning models over the past decade. In real-world practice, rule-based systems have their inherited characteristics—such as transparency, interpretability, and controlled adaptability—that remain challenges to replace with sophisticated machine learning approaches. However, building and maintaining such systems is labor-intensive, particularly for tasks involving a large variety of language expressions, where handcrafted rules must account for significant linguistic variability.

Although recent advancements in large language models (LLMs) have demonstrated exceptional performance across various NLP tasks with minimum labeled data for training, their high computational demands limit widespread adoption. LLMs leverage vast pretraining on extensive datasets to embed prior knowledge and semantic nuances, thereby reducing the volume of task-specific training data required for effective deployment.[2–5] Nonetheless, executing LLMs within a HIPAA-compliant environment requires significant computational resources and specialized expertise—resources that still challenge many healthcare organizations. Moreover, operationalizing LLM-based solutions is notably expensive, especially when dealing with large-scale datasets.

Previous studies explored various methods of hybridizing rule-based and machine-learning approaches to leverage the advantages of both and mitigate their respective shortcomings.[6–8] However, these hybrid approaches still did not address the transparency and interpretability issues brought by the machine learning models. The inherent uncertainty and randomness associated with machine learning-based models remain a concern to clinical users despite their probabilistic foundations.[9,10]

In light of these challenges, we proposed a novel approach that applies LLMs to assist in the development of rule-based NLP systems. While prior studies have reported the success of using LLMs to generate rules or regex, to our knowledge, no study has specifically investigated this direction within the clinical NLP domain. This approach offers promising advantages: although running LLMs can be resource-intensive, our proposed method employs them solely during the development phase. Once the development is complete, we can fully benefit from the efficiency and transparency of the finalized rule-based systems.

To demonstrate the feasibility of this proposed approach, we applied it to our previously developed rule-based NLP system, EasyCIE, for detecting surgical site infection (SSI).[11] This system was developed in accordance with the American College of Surgeons National Surgical Quality Improvement Program (NSQIP) guideline. It is now advancing to the operational phase for post-surgical complication surveillance, with the goal of enhancing patient outcomes and surgical safety. We selected this system for several reasons. First, its versatile architecture has been demonstrated to be effective across multiple clinical NLP tasks.[12,13] Second, its optimized rule-processing engine ensures high processing speed, even as the number of rules increases[14]—an important consideration when exploring automated rule generation using LLMs. Third, this pipeline is designed to decompose complex NLP tasks into simpler, modular subtasks, simplifying development and reducing the overall challenge.[15] Finally, the availability of an existing annotated dataset and established rules enables efficient evaluation of our approach. As an initial exploratory study, we investigated two focused hypotheses:

*Hypothesis 1*: Given limited example data, LLMs can identify clinically relevant snippets from patient notes that are useful for downstream NLP development.

*Hypothesis 2*: Once a snippet is identified, LLMs can accurately extract the key terms necessary for constructing the rules of the NER component.

**Methods**

Data Source

This study used the clinical data in the University of Utah Health system spanning from September 1, 2015, through December 31, 2018. The institution adopted the Epic electronic medical records (EMR) in 2015. The institutional review board approved the study with a waiver of informed consent for the use of patient data. Clinical notes were queried from the University of Utah Health enterprise data warehouse using NSQIP case identifiers. These data were previously reviewed by NSQIP-trained surgical clinical reviewers (SCR). We used the SSI definition of the NSQIP 2017 operations manual as the reference standard. A total of 21,784 cases were annotated at both snippet level — evidences supporting or refuting an SSI diagnosis and episode level refuting—whether the corresponding surgical procedure followed by an SSI event. Two trained annotators with clinical background used eHOST to complete the annotation task with partial overlapping, and disagreements were adjudicated by a third reviewer.[11]

Tasks formulation

The annotated data were parsed and segmented into sentences using medspaCy.[16] Because minor segmentation errors occurred (with a few cases producing multiple or partial sentences), the segmented text pieces were referred to as "snippets." For each snippet, we constructed a binary label indicating whether a snippet contained any annotation. Therefore, evaluating the LLMs' performance on this binary classification task can address the first hypothesis. Next, we processed these snippets using our previously developed NLP pipeline with a subset of NER rules and extracted all the snippets that produced any output from the NER component. These snippets were then used as the input for the second task: verifying if the keyword set derived by the LLMs can adequately cover the rule set used in the existing NER component, thereby testing our second hypothesis. These snippets were split into training and test sets.

LLMs solution development

The experiments were conducted within a HIPAA-compliant environment at the Center for High-Performance Computing at the University of Utah, using a single machine equipped with two A100 GPUs. Two instruct-tuned quantized LLMs were used: Deepseek R1 distilled Qwen 32B (Deepseek) and Qwen2.5 Coder 32B (Qwen). To iteratively refine the LLM prompts, we adopted the chain of thoughts (CoT)[17] and the mixture of prompt experts (MoPE)[18] strategies with few-shot examples. The training set was used to verify and refine the prompts. To enhance the method's generalizability, the NSQIP manual and annotation guideline were directly converted into markdown-formatted text at the beginning of prompts.

For each task, we followed two-step process: a reasoning step and a verification step. In the reasoning step, the NSQIP manual and instructions requested the models to walkthrough the task step by step. In the verification step, the annotation guideline and additional prompts were used to instruct the models to validate the output from the reasoning step and generate final conclusions, including formatted output. The formatted output will be parsed and used for evaluation. For each case in task 1, each case was processed five times, with the final prediction determined by majority vote. The detailed prompts were provided in the appendix.

Evaluation

The test set was used for evaluation. For the snippet identification task, we measured the LLMs' performance using precision, recall, and F1 scores. For the keyword extraction task, we measured the percentage of the snippets that generated keywords can cover compared with original rules. Because both steps represent the initial steps of the whole NLP workflow, our primary focus was on recall rather than precision. In the second task, we did not evaluate precision, as the NER component's impact on final performance is non-deterministic alone; precisions can be further refined in collaboration with downstream components, such as context detector.

## Results

Table 1 shows the snippet identification performance of the Deepseek and Qwen models. Both achieved impressive recall with reasonable precision. We further analyzed the errors by sampling 100 errors from each model's outputs, with the results summarized in Table 2. As the analysis revealed, 15 out of 100 errors for Deepseek and 24 out of 100 errors for Qwen were true false positives, while the remaining "errors" provided varying degrees of informative cues for potential SSI diagnosis. More details were discussed in the next section. In the keyword extraction task, both Qwen and Deepseek model generated keywords achieved 100% snippets coverage.

**Table 1.** The performance of snippet identification task.

| Models | Precision | Recall | F1 score |
|---|---|---|---|
| Deepseek | 0.10 | 0.98 | 0.18 |
| Qwen | 0.08 | 0.99 | 0.15 |

**Table 2.** False positive error analyses of snippet identification task.

| Type of Errors | Counts | |
|---|---|---|
| | **Deepseek** | **Qwen** |
| Abnormal lab tests indicating infection | 4 | 4 |
| Anatomic site | 7 | 6 |
| Date/time information that can be used to determine SSI | 1 | 2 |
| Evidence that highly suggest an SSI or local infection | 9 | 0 |
| External tube access | 12 | 6 |
| General context without direct SSI relevance** | 7 | 7 |
| Infection | 1 | 4 |
| Invalid output format* | 0 | 9 |
| Irrelevant snippet* | **15** | **24** |
| Risk factor that can contribute to SSI | 18 | 12 |
| Site infection on the surgery date | 1 | 0 |
| Surgical procedure | 9 | 16 |
| Symptoms that can be caused by SSI | 11 | 6 |
| Treatment for infection or symptoms | 8 | 7 |

**Note**: Some snippets fall into multiple types. * This category reflects the true false positives. ** This category is debatable (see details in the discussion section below)

## Discussion

The two central hypotheses guided our study to provide an initial assessment regarding the feasibility of using LLMs to assist rule-based clinical NLP development. These two hypotheses covered two aspects: the ability of LLMs to identify clinically relevant snippets and the capability to extract key terms for NER rule construction. Our evaluation of snippet identification demonstrated that both Deepseek and Qwen models were capable of capturing informative snippets with exceptionally high recall (Deepseek: 0.98; Qwen: 0.99). This high recall ensured all relevant snippets were gathered for the downstream development process. Despite the notable low precision, our error analyses revealed that many of these "false positive" cases actually contained useful cues that could support the determination of SSI. This is because: 1. during the annotation phase, annotators focused on the document-level and episode-level labels, where they tended not to annotate duplicated snippet-level cues, 2. If a note is SSI irrelevant, concepts like infection without specific anatomic location or antibiotic treatment without explicit mention of SSI were tended not to be annotated. In the keyword extraction task, both models achieved very high coverage. These findings provide preliminary support for our approach while highlighting areas for optimization and integration with downstream components.

*Snippet identification error analysis*

The false negative errors were minimal (8 by Deepseek, 3 by Qwen), partly due to our instructions emphasizing the importance of including not only direct evidence but also factors that can contribute to potential SSI (See appendix). However, these errors seemed to reflect the models' limitations. Both models overlooked an ill-formatted snippet, which included a string converted from a lab report table followed by a sentence describing wound infection. When isolating the wound infection sentence, both models predicted correctly, though not when processing the original text. The remaining errors were inexplicably unrecognized by models. For example, "Of note, he was also reported to have a possible pneumonia on xx/x/xx and was taking amoxicillin intermittently. He was taken to the OR and had a perforated bowel that required 35cm of ileum resected." This snippet indicated an infection at the time of surgery. Although it cannot determine if an SSI occurred afterward, it certainly should have been included but was missed by Qwen, and similar snippets missed by Deepseek. These errors suggested either inherent limitations in the models or shortcomings in the prompt engineering for these two models.

The error analyses revealed that most of the false positively identified snippets were, in fact, clinically useful for determining SSI. In Table 2, we summarized 14 error types, most of which stemmed not from the unexpected behavior or mistakes of LLMs but from the difference between annotators' judgments and the strict adherence of LLMs to the instructions. When annotators reviewed the data, they considered the broader clinical context, often skipping snippets that did not clearly support or refute a potential SSI diagnosis. In contrast, the LLMs strictly evaluated snippets based solely on their explicit content, without broader context, resulting in apparent discrepancies between annotators' judgments and models' outputs.

Several error categories illustrated this contextual discrepancy clearly. For example, snippets classified under "Abnormal lab tests indicating infection," such as elevated white blood cell counts, were often considered non-informative when viewed in isolation or in notes without any other evidence of infection. However, in notes where an SSI is suspected, such findings can contribute to a higher probability of an SSI diagnosis. Similar patterns appeared in the "Infection" category, where mentions of other infections or infections without specifying sources (e.g., "# MRSA bacteremia # HCAP") could help interpret ambiguous symptoms like fever in determining SSI likelihood. Another relevant category was "Symptoms that can be caused by SSI" and "Infection treatment," where snippets described wound-care treatments or symptomatic interventions, indirectly hinting at underlying infection risk or management. Including these snippets in the first place can be valuable for downstream analyses. However, when annotators observed that they were not linked to a potential SSI, they were often not annotated.

Several other categories highlighted how the models' strict snippet-focused assessment in situations requiring broader context clinical interpretation. For instance, snippets categorized as "Anatomic site", such as "Right groin incision healing well," provided critical site-specific information that could rule out superficial SSIs. However, if if such a snippet were merely part of a routine follow-up after a thoracic surgery, it could be considered irrelevant. Similarly, the category "Date/time information" included snippets like admission dates, which is valuable in determining the timing and classification of SSI (e.g., differentiating Present At the Time Of Surgery—PATOS). "External tube access" snippets, such as "Monitor incisions and JP site for s/s of bleeding or infection," contributed useful anatomical site references and indicated heightened SSI risks despite not directly confirming SSI. Likewise, the "Surgical procedure" category contained essential procedural information to determine if a later local infection was on the same location.

Thirdly, a small number of "errors" were, in fact, likely missed by annotators, including "Site infection on surgery date (PATOS)" and "Evidence that highly suggests an SSI or local infection" were overlooked. Specifically, the statement "Operative course was complicated by findings on peritoneal exploration that the appendix was diffusely inflamed" (first category) and "To OR for left groin washout and sartorius flap mobilization" (second category) were not annotated.

Next, a small set of errors were either genuinely irrelevant or reflected the technical limitations of the models. The "Irrelevant snippets" were the ones where models often struggled. For example, consider the snippet: "Constitutional: Well nourished, well developed, Moderate distress." In one instance, Deepseek argued that "the snippet should be collected as it provides a symptom (distress) that could be part of the broader assessment for an SSI, even though it lacks specific details." However, in another instance, Deepseek reasoned that "The clinical note snippet provided does not contain specific information that would help in diagnosing a Surgical Site Infection (SSI). It describes the patient's general condition and level of distress but lacks details about signs or symptoms directly related to a surgical site infection." Similar struggles were observed in the category of "General context without direct SSI relevance." For instance, "SKIN: no worrisome lesions, no rashes" can be useful to refute a superficial SSI, if within a context of description surgical site, however, not informative by itself in general cases. Additionally, the category "Invalid output format" indicated technical limitations, notably more frequent in the Qwen model than in Deepseek, where responses failed to follow the required output formatting. Future optimization, possibly through larger or non-quantized models, could mitigate such errors.

Lastly, the category "Risk factor that can contribute to SSI" is one area where errors could have been reduced, yet we deliberately chose not to. In our original prompt instructions, we did not instruct the models to identify risk factors potentially contributing to SSI. However, since the NSQIP manual was directly incorporated into the prompts, the models tended to prioritize the immediate clinical interpretation related to diagnosing SSI, occasionally overlooking our explicit direction to first capture all potentially useful information. Consequently, the models missed some valuable cues, such as "perforated intestine," which could significantly inform SSI assessments. Conversely, after we added the instructions, other genuine risk factors, such as chemotherapy or diabetes, were appropriately included by the models. These inclusions, although technically flagged as errors, were still beneficial for SSI diagnosis, particularly when direct evidence was limited or inconclusive, and thus should not be considered true false positives.

*Findings of the keyword extraction task*

Not surprisingly, both models excessively generate keywords that covered the snippets of the original rules can match. However, we observed a wide range of flexibility in how these keywords can be generated. For example, given the snippet "Moderate volume abdominal and pelvic ascites, with diffuse thickening and enhancement of the peritoneal surfaces," the original rules included only "peritoneal" and "abdominal" as anatomical sites. Without explicitly instructing the models to avoid overgeneralization (which would result in numerous false positives), generated keywords often included broader terms like "thickening" and "enhancement." After adding constraints to the instructions, the models produced more conservative and precise keywords, such as "abdominal and pelvic ascites," "diffuse thickening," and "enhancement of the peritoneal surfaces." Additionally, the expanded keywords instructions seem inspiring to make the rules more generalizable, such as: "abdominal ascites," "pelvic ascites," "peritoneal enhancement," and "thickened peritoneum." Additionally, the expanded keyword suggestions inspired more generalizable rules, such as "abdominal ascites," "pelvic ascites," "peritoneal enhancement," and "thickened peritoneum." With these valuable candidate keywords generated by the models, NLP developers can create rules more efficiently and with less labeled data.

*Limitations*

The current findings of this study are specific to the SSI detection task and dataset used and may not generalize directly to other clinical domains or institutions without further validation. We did not exhaustively engineer the prompts, tested on larger models, or experimented different models' parameters, e.g. temperature. These aspects provide opportunities for future improvement.

Admittedly, this study represents only an initial exploration of LLM capabilities in supporting rule-based NLP development. Although the snippet identification performance surpassed our expectations with excellent sensitivity and better-than-expected precision, the keywords generated currently remain distant from the optimized rule set utilized in the NER component. Furthermore, our existing EasyCIE NER component supports PSEUDO rules—rules that allow PSEUDO-matched concepts to override other matched concepts, thereby reducing false-positive matches. However, we did not evaluate prompts for generating this specific type of rule in this study. We believe that employing reinforcement learning informed by feedback from rule execution could further enhance model performance for this task, and we intend to investigate this promising approach in future research.

**Conclusion**
This study explored a novel approach of using LLMs to assist clinical rule-based NLP system development. The results demonstrated that with properly constructed prompts, LLMs are capable of identifying relevant text snippets and suggesting meaningful keyword candidates for NER rule development. Starting from this first step toward more efficient, cost-effective, and transparent NLP solutions compared to purely deep-learning-based methods, we plan to work on refining prompt engineering, utilizing reinforcement learning for improved rule generation, extending the work to downstream NLP component development, and testing generalizability across diverse clinical tasks.

**Acknowledgement**
The support and resources from the Center for High Performance Computing at the University of Utah are gratefully acknowledged. The computational resources used were partially funded by the NIH Shared Instrumentation Grant 1S10OD021644-01A1.

17. Wei, J. *et al.* Chain-of-Thought Prompting Elicits Reasoning in Large Language Models. Preprint at https://doi.org/10.48550/arXiv.2201.11903 (2023).
18. Jiang, R., Liu, L. & Chen, C. MoPE: Mixture of Prompt Experts for Parameter-Efficient and Scalable Multimodal Fusion. Preprint at https://doi.org/10.48550/arXiv.2403.10568 (2025).

**Appendix**

Note: the few shot examples are omitted here because of page limit.

*Snippet identification prompts:*

**Reasoning prompt:**

{SSI guideline}

-------------------------------

# Your Role: Surgeon

# Task: Collect useful information for diagnosing SSI

You need to collect all useful information for diagnosing SSI. Given the limited context provided by the clinical note snippet, determine whether the snippet itself should be collected.

Even if the snippet alone is insufficient (e.g., missing anatomic location, surgical site), it can be informative when additional information is provided; consider the snippet useful and include it.

-------------------------------

## Given the input snippet:

{text}

## Walk through the reasoning process of the following:

{subtask}

### Reasoning:

**The {subtask} in the above prompt includes two subtask prompts**:

**Signs or Symptoms**

   - Analyze the snippet for any mention of signs or symptoms that might be related to a current or recent infection or any potential infection, even if its possibility is low.

   - Analyze the snippet for any mention of signs or symptoms that could increase the risk of SSI, which are potentially useful for diagnosing PATOS.

   - Exclude infections that are not possible of SSI, e.g. pneumonia. Do not exclude infections without any specific locations or reasons.'")

**Treatment Information**

   - Identify any references to treatments (e.g., planned, ongoing, recently completed, or stopped) that could be associated with managing or addressing infection.

- Exclude any treatments that are explicitly documented for other infection than SSI. Do not exclude treatment for the infection without any specific locations or reasons.

**Verification prompt**:

{SSI annotation guideline}

As a surgeon, you review a clinical note snippet and another surgeon's opinion regarding whether the snippet provides any useful information for diagnosing surgical site infection and determine whether the snippet itself should be collected.

**Note**: Even if a snippet by itself is insufficient for diagnosis (e.g., if it is missing the anatomical location or surgical site), if it contributes any useful diagnostic information and other essential details are provided elsewhere, consider the snippet useful and include it. For example, if a snippet mentions infection without specifying an anatomical site, but a preceding snippet (not within the current snippet) already refers to an abdominal surgical wound, the two together can support a diagnosis of SSI. Thus, even when viewed in isolation, the 'infection' snippet should be collected for downstream analyses, and so should be the 'wound' snippet.

{text}

## Here are the tasks you need to complete:

1. Verify the opinion expressed by another surgeon. Is it valid or not?

2. Do you agree with the other surgeon's opinion? Why?

3. Summarize your own opinion which will be the final decision.

4. Generate a json-format output: either {{"conclusion":"yes"}} or {{"conclusion":"no"}}

*Keywords extraction prompts:*

Reasoning prompt:

{SSI annotation guideline}
--------------------------------
# Your Role: Clinical Informatist
# Task: Identify the keywords from the given snippet to be used in a rule-based NER component.
The following snippet contains useful information to help decide if a patient has SSI, either because it contains a piece of information that can be linked to SSI, or a piece of
information that can be linked to other infections to refute an SSI, or a piece of information that describe normality of a potential surgical site to refute an SSI. Hence, given the limited context provided by the clinical note snippet, even if the snippet alone is insufficient (e.g. an anatomic location without procedure, a surgical site without infection mention, or infection signs and symptoms without location), but it can be still informative
when additional information outside the given snippet is provided.
You are building a rule-based NER component to identify the above relevant information. Following the steps below, you need to identify the keywords directly from the snippet and explain how they can be relevant:
1. Identify all anatomic sites or body parts or mentions indicating anatomic sites that potentially can be a surgical site (can potentially be used to link a procedure or infection mentioned outside the given snippet)
2. Identify all surgical or invasive procedures (can potentially be used to link an infection mentioned outside the given snippet)
3. Identify all infection or infection treatment
4. Identify all wound care

5. Identify all other infections that can be used to explain signs, symptoms, or lab abnormalities that might or might not be caused by SSI.
6. Make sure the keywords can actually be found in the original text and capture the key information that can be used for downstream analysis. Also, you need to double-check to make sure they keywords themselves are not over-generalized to get massive matches in other notes.
7. Based on your clinical knowledge, expand the keywords with synonyms, so that same meaning in a different snippet can also be found.
8. Prune the keywords to remove certainty, temporality, and experiencer modifiers, e.g. 'concerning for,' as these modifiers will be identified in downstream NLP components.
9. Summarize your findings in a json dictonary using template: {{"concepts": [], "expanded_concepts":[]}}
-------------------------------
# For example:
### Given the input snippet:
She underwent CT scan of the abdomen that showed an appendix dilated to 11 mm with adjacent fat stranding and inflammation.

### Analysis:
1. **Anatomic Sites or Body Parts (Surgical Sites):**
   - **Appendix:** This is a clear anatomic site that could be a surgical site in the context of appendectomy.
2. **Surgical or Invasive Procedures:**
   - **CT scan:** This is an imaging procedure, but it is not an invasive procedure. It is more relevant for identifying anatomical findings and conditions rather than surgical interventions.
3. **Infection or Infection Treatment:**
   - **Inflammation:** This term indicates the presence of an inflammatory response, which could be related to an infection. However, it is not specific to surgical site infections (SSIs).
4. **Wound Care:**
   - **Not mentioned:** There is no mention of wound care in the snippet.
5. **Other Infections:**
   - **Inflammation:** This can be a sign of other infections, not specifically SSI. It could be related to a non-surgical infection, such as appendicitis.
6. **Keywords in Original Text:**
   - **Appendix, inflammation, CT scan:** These are the key terms that can be used for downstream analysis. They are specific enough to capture relevant information without being overly generalized.
7. **Expanded Concepts with Synonyms:**
   - **Appendix:** Appendix, cecum, right lower quadrant
   - **Inflammation:** Inflammation, inflammatory response
   - **CT scan:** CT scan, computed tomography
8. **Pruned Keywords:**
   - Remove any certainty, temporality, or experiencer modifiers. The terms are already quite specific.
### Summary in JSON Dictionary:
```json
{{
  "concepts": ["Appendix", "Inflammation", "CT scan"],
  "expanded_concepts": ["Appendix", "Cecum", "Right lower quadrant", "Inflammation", "Inflammatory response", "CT scan", "Computed tomography"]
}}
```
Now it's your turn to process the snippets.
### Given the input snippet:\n{text}
### Analysis:'

Verification prompt:
# Your Role: Surgeon
# Task: Given clinical note snippet and key terms (for rule-based NER) analysis, verify the analysis and confirm/revise the identified keywords to be used in a rule-based NER component.

The given snippet contains useful information to help decide if a patient has SSI. The goal is to build a rule-based NER component for this patient classification pipeline. Here are the steps to achieve the goal:
1. Identify the keywords directly from the snippet, make sure the keywords can actually found in original text and capture the key information that can be used to determine if a patient has SSI, even if the snippet doesn't have complete information. Also, you need to double check to make sure the keywords themselves are not over generalized to get massive matches in other notes.
2. Based on your clinical knowledge, expand the keywords with synonyms, so that same meaning in a different snippet can also be found.
3. Prune the keywords to remove certainty, temporality, and experiencer modifiers, e.g. 'concerning for', as these modifiers will be identified in downstream NLP components.
3. Elaborate with detailed justification
4. Summarize your findings in a json dictionary.

# Your task:
You will be given the clinical note snippet and the analysis from anther clinical informatist:
1. Verify the analysis to check if it follows the above instructions step by step. Pay attention to the expanded keywords, make sure none of them is too broad.
2. Double check the derived keywords list in json format.
3. Finalize the keywords in json format using the same structure: {{"concepts": [], "expanded_concepts":[]}}

{text}